# FMMI: Flow Matching Mutual Information Estimation

**Ivan Butakov** [*1 2] **Alexander Semenenko** [*1] **Valeriya Kirova** [3] **Alexey Frolov** [1] **Ivan Oseledets** [1 2]

## Abstract

We introduce a novel Mutual Information (MI) estimator that fundamentally reframes the discriminative approach. Instead of training a classifier to discriminate between joint and marginal distributions, we learn a normalizing flow that transforms one into the other. This technique produces a computationally efficient and precise MI estimate that scales well to high dimensions and across a wide range of ground-truth MI values.

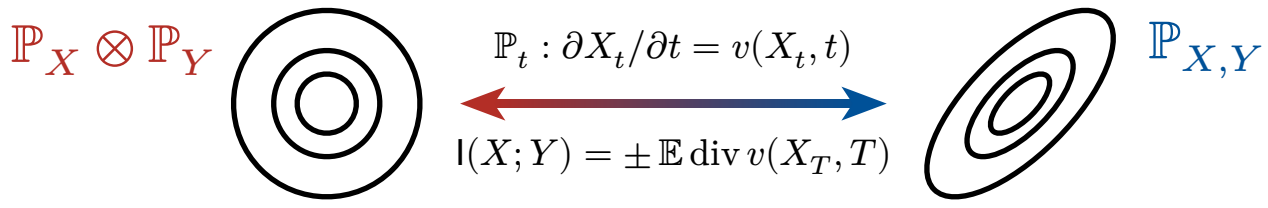

Code available at `pytorch-fmmi`

## 1. Introduction

Mutual Information (MI) is a fundamental measure of non-linear statistical dependence between two random vectors, defined as the Kullback-Leibler divergence between the joint distribution and the product of marginals (Polyanskiy and Wu, 2024):

$$\mathsf{I}(X;Y) = \mathsf{KL}\left[\mathbb{P}_{X,Y} \,\middle\|\, \mathbb{P}_X \otimes \mathbb{P}_Y\right]$$

It is well-defined for most joint distributions, non-negative and zero if and only if $X$ and $Y$ are independent, invariant to bijections, and possesses many other useful properties.

For these reasons, MI is used extensively for theoretical analysis of overfitting (Asadi et al., 2018; Negrea et al., 2019), hypothesis testing (Duong and Nguyen, 2022), feature selection (Battiti, 1994; Peng et al., 2005; Vergara and Estévez, 2014) representation learning (Bachman et al., 2019; Butakov et al., 2025; Hjelm et al., 2019; Tschannen et al., 2020; Veličković et al., 2019), and studying mechanisms behind generalization in deep neural networks (DNNs) (Butakov et al., 2024a; Goldfeld et al., 2019; Shwartz-Ziv and Tishby, 2017; Tishby and Zaslavsky, 2015).

In practical scenarios, $\mathbb{P}_{X,Y}$ and $\mathbb{P}_X \otimes \mathbb{P}_Y$ are unknown, requiring MI to be estimated from finite samples. This reliance on empirical estimates leads to the curse of dimensionality: the sample complexity of MI grows exponentially with the number of dimensions (Goldfeld et al., 2020; McAllester and Stratos, 2020). Long-tailed distributions and large values of MI further complicate the estimation (Czyż et al., 2023; McAllester and Stratos, 2020). These problems considerably limit the applications of information theory to real-scale problems. However, recent advances in the neural estimation methods show that complex parametric estimators achieve relative practical success in the cases where classical MI estimation techniques fail.

Contemporary parametric MI estimators fall into two broad categories: *discriminative* and *generative* (Song and Ermon, 2020). While the latter learn $\mathbb{P}_{X,Y}$ and $\mathbb{P}_X \otimes \mathbb{P}_Y$ from scratch, the former employ a classifier to discern between these two distributions. Despite the discriminative approach being more elegant, modern state-of-the-art MI estimators are mostly generative (Butakov et al., 2024b; Franzese et al., 2024; Kholkin et al., 2026). This reflects the inherent limitations of current discriminative methods: they are essentially *generalized energy-based models* (GEBMs) (Arbel et al., 2021), which are notoriously suboptimal for significantly different prior and posterior distributions.

In this work, we take a leap forward and leverage **continuous-time Normalizing Flows (CNFs)** and **Flow Matching (FM)** to advance the discriminative approach beyond GEBMs. Contrary to the traditional generative approach, we learn a function which transforms a product marginal distributions into a joint distribution (*couples* the data). It can be shown that, in continuous case, the expected log-Jacobian of this transform is precisely the Mutual Information between the coupled data. The flow matching technique is used to avoid the reliance on unknown data distribution. This approach can also be generalized to any multi-variable extension of MI.

Overall, our contribution is as follows:

1. We propose a *novel* and *universal* flow-based estimator of information quantities (**FMDoE** & **FMMI**).
2. We provide theoretical guarantees for our method and amortize it to reduce the computational load.

[*]Equal contribution [1]Applied AI Institute, Moscow, Russia [2]Institute of Numerical Mathematics, RAS, Moscow, Russia [3]National Research Nuclear University MEPhI, Moscow, Russia. Correspondence to: Ivan Butakov <ivan.butakov@applied-ai.ru>, Alexander Semenenko <semenenko@applied-ai.ru>.

3. The estimator is evaluated across a diverse set of benchmarks, including high-dimensional, high-MI and other challenging setups. The results indicate the superiority of our method.

The remainder of the paper is structured as follows: in Section 2, the necessary background in information theory and flow models is provided; the related works are discussed in Section 3; Section 4 introduces a general approach that is then reduced to the MI and O-information estimation in Section 5 and Section 6; the method is evaluated across numerous setups and competitors in Section 7; finally, we discuss the results in Section 8.

## 2. Background

**Information Theory.** Let $(\Omega, \mathcal{F}, \mathbb{P})$ be a probability space with sample space $\Omega$, $\sigma$-algebra $\mathcal{F}$, and probability measure $\mathbb{P}$ defined on $\mathcal{F}$. For another probability measure $\mathbb{Q}$ with $\mathbb{Q} \ll \mathbb{P}$, the Kullback-Leibler (KL) divergence is $\mathsf{KL}[\mathbb{Q} \parallel \mathbb{P}] = \mathbb{E}_{\mathbb{Q}}\left[\log \frac{\mathrm{d}\,\mathbb{Q}}{\mathrm{d}\,\mathbb{P}}\right]$, which is non-negative and vanishes if and only if (iff) $\mathbb{P} = \mathbb{Q}$.

Consider random vectors $X : \Omega \to \mathcal{X}$ and $Y : \Omega \to \mathcal{Y}$ with joint distribution $\mathbb{P}_{X,Y}$ and marginals $\mathbb{P}_X$ and $\mathbb{P}_Y$, respectively. We denote product measures by $\mathbb{P}_X \otimes \mathbb{P}_Y$. Wherever needed, we assume the relevant Radon-Nikodym derivatives exist. The mutual information (MI) between $X$ and $Y$ quantifies the divergence between the joint distribution and the product of marginals:

$$\mathsf{I}(X;Y) = \mathbb{E} \log \frac{\mathrm{d}\,\mathbb{P}_{X,Y}}{\mathrm{d}\,\mathbb{P}_X \otimes \mathbb{P}_Y} = \mathsf{KL}\left[\mathbb{P}_{X,Y} \parallel \mathbb{P}_X \otimes \mathbb{P}_Y\right] \quad (1)$$

MI posses several outstanding properties. Namely, $\mathsf{I}(X;Y) = 0$ iff $X \perp\!\!\!\perp Y$ and $\mathsf{I}(X;Y) = \mathsf{I}(X; g(Y))$ for an invertible and measurable $g$ (Polyanskiy and Wu, 2024).

When $\mathbb{P}_X$ admits a probability density function (PDF) $p_X$ with respect to (w.r.t.) the Lebesgue measure, the differential entropy is defined as $\mathsf{h}(X) = -\mathbb{E}[\log p_X(X)]$, where $\log(\cdot)$ denotes the natural logarithm. Likewise, the joint entropy $\mathsf{h}(X,Y)$ is defined via the joint density $p_{X,Y}(x,y)$, and conditional entropy is $\mathsf{h}(X \mid Y) = -\,\mathbb{E}\left[\log p_{X|Y}(X \mid Y)\right]$. Under the existence of PDFs, MI satisfies the identities

$$\begin{aligned} \mathsf{I}(X;Y) &= \mathsf{h}(X) - \mathsf{h}(X \mid Y) \\ &= \mathsf{h}(Y) - \mathsf{h}(Y \mid X) \\ &= \mathsf{h}(X) + \mathsf{h}(Y) - \mathsf{h}(X,Y). \end{aligned} \quad (2)$$

**Optimal Transport.** Consider a Polish metric space $(M, d)$. Let $\Gamma(\mathbb{P}_X, \mathbb{P}_Y)$ be a set of joint measures $\mathbb{P}_{X,Y}$ (*couplings*) whose marginals are $\mathbb{P}_X$ and $\mathbb{P}_Y$, defined on $M$. For $p \geq 1$, the Wasserstein $p$-distance between $\mathbb{P}_X, \mathbb{P}_Y$ is

$$\mathsf{W}_p(\mathbb{P}_X, \mathbb{P}_Y) = \inf_{\mathbb{P}_{X,Y} \in \Gamma(\mathbb{P}_X, \mathbb{P}_Y)} \sqrt[p]{\mathbb{E}\, d(X,Y)^p}$$

Following the idea behind (1), one can define the *Wasserstein dependency measure* (WMI) (Ozair et al., 2019):

$$\mathsf{WI}_p(X;Y) = \mathsf{W}_p\big(\mathbb{P}_{X,Y}, \mathbb{P}_X \otimes \mathbb{P}_Y\big)$$

While this measure maintains some key properties of conventional MI (namely, nullification iff $X \perp\!\!\!\perp Y$), it also loses many of them (e.g., the invariance to bijections and DPI). Hence it is not as widely adopted as MI.

**Normalizing Flows.** Consider two absolutely continuous distributions $\mathbb{P}$ and $\mathbb{Q}$ defined on $\mathbb{R}^d$ with PDFs $p$ and $q$ correspondingly. Suppose we want to learn an invertible transform $f : \mathbb{R}^d \to \mathbb{R}^d$ such that $\mathbb{Q}$ is a push-forward of $\mathbb{P}$:

$$\mathbb{P} = \mathbb{Q} \circ f, \quad p(x) = q(f(x)) \cdot \left|\det \frac{\partial f}{\partial x}\right|(x) \quad (3)$$

If $q(x)$ is tractable, $f$ can be learnt in a maximum-likelihood manner (Tabak and Turner, 2013):

$$\mathbb{E}_{X \sim \mathbb{P}}\left[\log q\big(\hat{f}(X)\big) + \log\left|\det \frac{\partial \hat{f}}{\partial x}\right|(X)\right] \to \max_{\hat{f}}$$

Most contemporary normalizing flows are *continuous* (Tabak and Vanden-Eijnden, 2010). That is, they are parametrized by a velocity field through the following ordinary differential equation (ODE):

$$\frac{\partial x_t}{\partial t} = v(x_t, t), \quad t \in [0;1], \quad f(x_0) \overset{\text{def}}{=} x_1(x_0) \quad (4)$$

In this case, the change of variables formula becomes[1]

$$\frac{\mathrm{d}}{\mathrm{d}t}[\log p_t(x_t)] = -\operatorname{div} v(x_t, t), \qquad \begin{aligned} p_0 &= p \\ p_1 &= q \end{aligned} \quad (5)$$

While calculating the divergence naïvely requires $O(d)$ differentiations, an unbiased Hutchinson trace estimator (Hutchinson, 1989) can be used to reduce the cost down to $O(1)$ (Grathwohl et al., 2019):

$$\operatorname{div} v(x,t) = \operatorname{tr} \frac{\partial v}{\partial x} = \mathbb{E}\, a^{\mathsf{T}} \frac{\partial v}{\partial x} a, \quad \text{where} \quad \begin{aligned} \mathbb{E}\, a &= 0 \\ \operatorname{cov}(a) &= \mathrm{I} \end{aligned}$$

**Flow Matching.** Vanilla CNFs possess *two severe limitations*: the maximum-likelihood training requires (a) $q$ to be tractable and (b) $x_t$ to be simulated through a numerical ODE integration for each data point.

Framing velocity field learning as a regression (*matching*) task alleviates these issues. Consider any tractable conditional velocity field $v(x, t \mid x_1)$ that transforms $p_0$ into delta-measure $\delta_{x_1}$ through a conditional probability path

[1] Hereinafter div and $\nabla$ are with respect to $x$, if not stated otherwise.

$p_{X_t|x_1}$ with tractable expectations. One can then learn $v(x, t \mid x_1)$ via the following objective:

$$\mathbb{E}\left\| v\Big(X_{T|x_1}, T \;\Big|\; x_1\Big) - \hat{v}\Big(X_{T|x_1}, T\Big) \right\|_2 \to \min_{\hat{v}}, \quad (6)$$

where $T \sim \mathrm{U}[0;1]$ and $X_{t|x_1} \sim p_{X_t|x_1}$. However, we are interested in learning the unconditional velocity field. Thankfully, averaging (6) over $x_1 \sim p_1$ produces a valid objective for matching $v(x,t)$ (Lipman et al., 2023):

$$\mathbb{E}\left\| v(X_T, T \mid X_1) - \hat{v}(X_T, T) \right\|_2 \to \min_{\hat{v}}, \quad (7)$$

where $X_1 \sim p_1$ and $X_t = X_{t|X_1}$. If (7) attains its minima, then $\hat{v}$ transforms $p_0$ to $p_t$ through (5) for any $t \in [0;1]$.

A common choice for $p_t$ and $v(x, t \mid x_1)$ is *linear interpolation* or *conditional optimal transport* (Lipman et al., 2023):

$$X_t = (1-t)X_0 + t\,X_1 \quad v(x, t \mid x_1) = x_1 - x, \quad (8)$$

where pairs $(X_0, X_1)$ are sampled from a coupling distribution (typically independent) with marginals $p_0$ and $p_1$. However, other options also exist (Lipman et al., 2024).

## 3. Related Works

Non-parametric methods pioneered the field by providing cheap MI estimates through binning (Moddemeijer, 1989), kernel (Moon et al., 1995) and $k$-nearest neighbors (Kozachenko and Leonenko, 1987; Kraskov et al., 2004) density estimation. However, recent studies show that these approaches fail horribly on high-dimensional and high-MI tasks (Butakov et al., 2024b; Czyż et al., 2023). While random slicing scales these methods to higher dimensions (Goldfeld and Greenewald, 2021), it also exhibits severe inherent limitations (Semenenko et al., 2026).

In contrast, complex parametric estimators have achieved relative practical success in dealing with difficult distributions. This family consist of *generative* and *discriminative* methods. While the generative estimators approximate $\mathbb{P}_{X,Y}$, $\mathbb{P}_X$ and $\mathbb{P}_Y$ from scratch, thus providing a plug-in estimate of MI (usually through Equation (2)), the discriminative approaches focus on the direct estimation of $\frac{\mathrm{d}\,\mathbb{P}_{X,Y}}{\mathrm{d}\,\mathbb{P}_X \otimes \mathbb{P}_Y}$ (Federici et al., 2023; Song and Ermon, 2020).

**Generative methods.** Naïve application of generative models to approximate densities typically produce highly biased MI estimates due to misalignment of estimated $\hat{\mathbb{P}}_{X,Y}$, $\hat{\mathbb{P}}_X$ and $\hat{\mathbb{P}}_Y$ (Song and Ermon, 2020). Subsequent research has addressed this issue by using a single model to estimate all three distributions simultaneously (Butakov et al., 2024b; Chen et al., 2025; Dahlke and Pacheco, 2025; Duong and Nguyen, 2023; Ni and Lotz, 2025). Among these methods, we highlight the **RFMI** estimator (Wang et al., 2025), which uses flow matching with a Gaussian prior to estimate $\mathbb{P}_X$ and $\mathbb{P}_{X|Y}$.

A separate subfamily of generative methods has recently emerged that avoids density estimation altogether, instead employing *diffusion models* and Girsanov's theorem (Franzese et al., 2024; Kholkin et al., 2026). While these methods provide better estimates, they demand orders of magnitude more samples and compute. These estimators also typically fail when faced with long-tailed distributions.

Overall, both families approximate the data distribution from scratch, thus performing unnecessary extra work.

**Discriminative methods.** While the direct estimation of density ratios is the most elegant approach, modern discriminative estimators face severe limitations. These include high demands on batch and sampling sizes (McAllester and Stratos, 2020; Oord et al., 2019), as well as constraints on the level of mutual information they can reliably measure (Poole et al., 2019; Song and Ermon, 2020). Since all contemporary discriminative methods are classifier-based, these issues mirror the limitations of generalized energy-based models (GEBMs) (Arbel et al., 2021). For instance, GEBMs struggle to estimate the density ratio between two highly dissimilar distributions (Rhodes et al., 2020), a scenario that corresponds to high MI in (1).

The literature primarily addresses these problems in two ways. One line of research involves alternative classification objectives (Letizia et al., 2024; Liao et al., 2020), which offer modest practical improvements but fail to overcome the fundamental theoretical limitations. The other focuses on telescoping density ratio estimation (Choi et al., 2022; Rhodes et al., 2020). This approach achieves significantly higher accuracy by training multiple (potentially infinite, as in **DRE-$\infty$** by Choi et al. (2022)) discriminators; however, it explicitly requires a path of distributions $\mathbb{P}_t$ with well-defined density ratios that interpolates between $\mathbb{P}_X \otimes \mathbb{P}_Y$ and $\mathbb{P}_{X,Y}$, which is not generally available out-of-the-box.

**Relation to our approach.** Our method is directly inspired by the **RFMI** and **DRE-$\infty$**, taking the best from the generative and discriminative approaches. Similarly to RFMI, we leverage flow matching to learn $\mathbb{P}_{X,Y}$ (or $\mathbb{P}_{X|Y}$), but using $\mathbb{P}_X \otimes \mathbb{P}_Y$ (or $\mathbb{P}_X$) instead of the Gaussian prior.

This connects us to DRE-$\infty$, which estimates the density ratio between two distributions in a continuous, bridge-like manner. However, instead of learning the density ratio evolution at any fixed point (which is often an ill-posed problem), we "follow" samples along the probability path, thus avoiding degenerate density ratios.

For additional discussion, please refer to Appendix B.

## 4. Core Method

Consider an absolutely continuous $X_0$ and a smooth bijective mapping $f$. Define $X_1 = f(X_0)$. Due to (3),

$$\mathsf{h}(X_1) = \mathsf{h}(X_0) + \mathbb{E} \log \left| \det \frac{\partial f}{\partial x}(X_0) \right|$$

Suppose $f$ is parametrized by a velocity field $v$. By (5),

$$\mathsf{h}(X_1) - \mathsf{h}(X_0) = \mathbb{E} \int_0^1 \operatorname{div} v(x_t(X_0), t) \, \mathrm{d}t$$

This equation requires solving ODE to simulate $x_t$. However, if expectations over $\mathbb{P}_{X_t}$ are tractable, a simulation-free result can be derived:

**Lemma 4.1.** For any $t \in [0; 1]$, let $X_t$ satisfy $\partial X_t / \partial t = v(X_t, t)$. Let $T \sim \mathrm{U}[0; 1]$. Then

$$\mathsf{h}(X_1) - \mathsf{h}(X_0) = \mathbb{E} \operatorname{div} v(X_T, T)$$

In practice, however, $f$ and $v$ are unknown, but independent samples from $p_0$ and $p_1$ are available. Since none of the true distributions are tractable, we propose using flow matching to learn $v$ from samples via (7) — see Algorithm 1. The estimation is Monte-Carlo-based — see Algorithm 2.

Note that estimated $\hat{v}$ typically does not produce the same probability path $\mathbb{P}_{X_t}$. To address this, we provide the following approximation and convergence analysis:

**Theorem 4.2.** Consider $\mathbb{P}_{X,T}$ such that $p_t(x) \overset{\text{def}}{=} p(x \mid t)$ exists, is smooth, compactly supported, and $\log p_t$ is $L_t$-Lipschitz in $x$. Then, for any smooth vector field $\epsilon(x, t)$,

$$|\mathbb{E} \operatorname{div} \epsilon(X, T)| \leq \sqrt{\mathbb{E} L_T^2 \cdot \mathbb{E} \|\epsilon(X, T)\|_2^2}$$

**Corollary 4.3.** (FMDoE approximation error) Under the setup from Lemma 4.1, define $\mathrm{DoE}_u = \mathbb{E} \operatorname{div} u(X_T, T)$ for any $u(x, t)$. For all $t \in [0; 1]$, let $p_t$ be smooth and compactly supported, $\log p_t$ be $L_t$-Lipschitz in $x$. Then

$$|\mathrm{DoE}_v - \mathrm{DoE}_{\hat{v}}| \leq \sqrt{\mathbb{E} L_T^2 \cdot \mathbb{E} \|v(X_T, T) - \hat{v}(X_T, T)\|_2^2}$$

*Proof.* Substitute $\epsilon = v - \hat{v}$ in Theorem 4.2. □

**Theorem 4.4.** (FMDoE convergence rate) Under the assumptions of Corollary 4.3 and Theorem 4.4 in (Zhou and Liu, 2025), suppose $v$ is $K$-Lipschitz in $x$ and $\operatorname{var}[\operatorname{div} v(X_T, T)] \leq \sigma^2$. Let $\hat{v}$ be the FM estimate from $N$ i.i.d. samples from $\mathbb{P}_{X_0, X_1}$, and $\widehat{\mathrm{DoE}}_{\hat{v}}$ be the MC estimate of $\mathrm{DoE}_{\hat{v}}$. Then, $\forall \delta \in (0, 1)$, with probability at least $1 - \frac{1}{N} - \delta$ over the random training and MC samples,

$$\left| \mathrm{DoE}_v - \widehat{\mathrm{DoE}}_{\hat{v}} \right| \leq \tilde{O}\left( \frac{\sqrt{\mathbb{E} L_T^2} \, K^{d/4}}{(1 - T)^2 N^{1/(d+5)}} \right) + O\left( \frac{\sigma}{\sqrt{\delta N}} \right),$$

where $\tilde{O}$ ignores logarithmic factors in $d$, $\log N$, and $\log(1 - T)$, and $f(x) = O(g(x))$ means $|f(x)| \leq |C g(x)|$ for some $C > 0$.

**Algorithm 1** FMDoE, training

1: **Input:** samples from $\mathbb{P}_{X_0, X_1}$, sampler from $\mathbb{P}_{X_t | X_1}$, initial velocity network $\hat{v}$.
2: **Output:** learned velocity network $\hat{v}$.
3: **while** not converged **do**
4: Sample batch of pairs $\{x_0^n, x_1^n\}_{n=1}^N \sim \mathbb{P}_{X_0, X_1}$
5: Sample batch $\{t^n\}_{n=1}^N \sim \mathrm{U}[0; 1]$
6: Using $\{x_0^n, x_1^n\}_{n=1}^N$, sample $\{x_t^n\}_{n=1}^N \sim \mathbb{P}_{X_t | X_1}$
7: $\mathcal{L}(\hat{v}) \leftarrow \frac{1}{N} \sum_{n=1}^N \|\hat{v}(x_t^n, t) - v(x_t^n, t \mid x_1^n)\|_2^2$
8: Update $\hat{v}$ using $\partial \mathcal{L} / \partial \hat{v}$
9: **end**

**Algorithm 2** FMDoE, estimation

1: **Input:** samples from $\mathbb{P}_{X_0, X_1}$, sampler from $\mathbb{P}_{X_t | X_1}$, velocity network $\hat{v}$.
2: **Output:** estimated $\mathsf{h}(X_1) - \mathsf{h}(X_0)$ and $\mathsf{W}_p(\mathbb{P}_0, \mathbb{P}_1)$.
3: Sample batch of pairs $\{x_0^n, x_1^n\}_{n=1}^N \sim \mathbb{P}_{X_0, X_1}$
4: Sample batch $\{t^n\}_{n=1}^N \sim \mathrm{U}[0; 1]$
5: Using $\{x_0^n, x_1^n\}_{n=1}^N$, sample $\{x_t^n\}_{n=1}^N \sim \mathbb{P}_{X_t | X_1}$
6: $\widehat{\mathrm{DoE}} \leftarrow \frac{1}{N} \sum_{n=1}^N \operatorname{div} \hat{v}(x_t^n, t)$
7: $\widehat{\mathsf{W}}_p \leftarrow \sqrt[p]{\frac{1}{N} \sum_{n=1}^N \|\hat{v}(x_t^n, t)\|_p^p}$

Interestingly, a similar theory can be derived for $\mathsf{W}_p(\mathbb{P}_0, \mathbb{P}_1)$ between the marginals of $X_0$ and $X_1$:

**Lemma 4.5.** (Benamou and Brenier, 2000, Proposition 1.1) For absolutely continuous $\mathbb{P}_0, \mathbb{P}_1$ define

$$V(\mathbb{P}_0, \mathbb{P}_1) = \{ v \mid v \text{ transforms } \mathbb{P}_0 \text{ into } \mathbb{P}_1 \text{ via } (5) \}$$

Then, if $T \sim \mathrm{U}[0; 1]$ and $\partial X_t / \partial t = v(X_t, t)$, $X_0 \sim \mathbb{P}_0$,

$$\mathsf{W}_p(\mathbb{P}_0, \mathbb{P}_1) = \inf_{v \in V(\mathbb{P}_0, \mathbb{P}_1)} \sqrt[p]{\mathbb{E} \|v(X_T, T)\|_p^p}$$

Notice the similarity between the expressions in Theorem 4.2 and Lemma 4.5. Since $\mathsf{h}(X_1) - \mathsf{h}(X_0)$ is invariant to the choice of $v \in V(\mathbb{P}_0, \mathbb{P}_1)$, one can select the latter to attain the infimum in Lemma 4.5, yielding the following:

**Corollary 4.6.** Suppose the assumptions of Theorem 4.2 hold and $L_t \leq L$ for some $L$ and every $t \in [0; 1]$. Then

$$|\mathsf{h}(X_1) - \mathsf{h}(X_0)| \leq L \cdot \mathsf{W}_2(\mathbb{P}_0, \mathbb{P}_1)$$

Therefore, we also suggest estimating $\mathbb{E} \|v(X_T, T)\|_p^p$ as a cheap surrogate that is connected to both $\mathsf{W}_p(\mathbb{P}_0, \mathbb{P}_1)$ and $\mathsf{h}(X_1) - \mathsf{h}(X_0)$ (see line 7 in Algorithm 2).

## 5. Mutual Information

*Figure 1.* Flow Matching Mutual Information estimation (joint).

Our method can be readily extended to Mutual Information estimation. Recall that, in continuous case, MI can be expressed through the difference of entropies via (2). Each formula in (2) can be approximated through the FMDoE estimator by constructing flows either between the joint distribution and the product of the marginals, or between a marginal and its corresponding conditional distribution.

**jFMMI.** Setting $\mathbb{P}_0 = \mathbb{P}_X \otimes \mathbb{P}_Y$ and $\mathbb{P}_1 = \mathbb{P}_{X,Y}$, we learn a velocity field $v : \mathcal{X} \times \mathcal{Y} \times [0;1] \to \mathcal{X} \times \mathcal{Y}$ that performs *forward* transformation $\mathbb{P}_0$ to $\mathbb{P}_1$. Alternatively, one may consider the *reverse* transformation by setting $\mathbb{P}_0 = \mathbb{P}_{X,Y}$ and $\mathbb{P}_1 = \mathbb{P}_X \otimes \mathbb{P}_Y$ to be the product of marginals. These two formulations are theoretically equivalent, though their practical behavior may differ slightly, as observed in our experiments. By Lemma 4.1, for $Z_0 \sim \mathbb{P}_0$, $Z_1 \sim \mathbb{P}_1$, and $T \sim \mathrm{U}[0;1]$, we have $\mathsf{h}(Z_1) - \mathsf{h}(Z_0) = \mathbb{E} \operatorname{div} v(Z_T, T)$, which is equivalent to

$$\mathsf{I}(X;Y) = \mathsf{h}(X) + \mathsf{h}(Y) - \mathsf{h}(X,Y) = -\mathbb{E} \operatorname{div} v(Z_T, T).$$

In practice, a batch from $\mathbb{P}_{X,Y}$ is converted into samples from $\mathbb{P}_X \otimes \mathbb{P}_Y$ by independently permuting the $x$'s and $y$'s coordinates. Training and estimation then follow Algorithm 1 and Algorithm 2, repsectively.

**cFMMI.** In the conditional approach, for each fixed $y$ we set $\mathbb{P}_0 = \mathbb{P}_X$, $\mathbb{P}_1 = \mathbb{P}_{X|Y=y}$ and learn $v : \mathcal{X} \times \mathcal{Y} \times [0,1] \to \mathcal{X}$ that transforms $\mathbb{P}_X$ to $\mathbb{P}_{X|Y=y}$. Applying Lemma 4.1 for a given $y$, we have $\mathsf{h}(X \mid Y) - \mathsf{h}(X) = \mathbb{E}_Y \, \mathbb{E}[\operatorname{div} v(X_T, Y, T) \mid Y] = \mathbb{E}[\operatorname{div} v(X_T, Y, T)]$. Thus,

$$\mathsf{I}(X;Y) = \mathsf{h}(X) - \mathsf{h}(X \mid Y) = -\mathbb{E} \operatorname{div} v(X_T, Y, T).$$

In practice, unconditional samples are obtained by shuffling only the $x$'s samples while keeping the corresponding $y$'s fixed. Training and Monte Carlo estimation similarly use algorithms from the previous section.

**Analysis.** Since jFMMI operates in $(d_X + d_Y)$-dimensional space, while cFMMI works in $d_X$ (or $d_Y$) dimensions, the latter can be more efficient in high-dimensional setups. These estimators also benefit from the Hutchinson trace estimator for efficient divergence computation. Under the assumptions of Corollary 4.3 and Theorem 4.2, both estimators are consistent, with approximation error bounded by the flow matching error and the Lipschitz constants of the true distributions.

## 6. O-Information

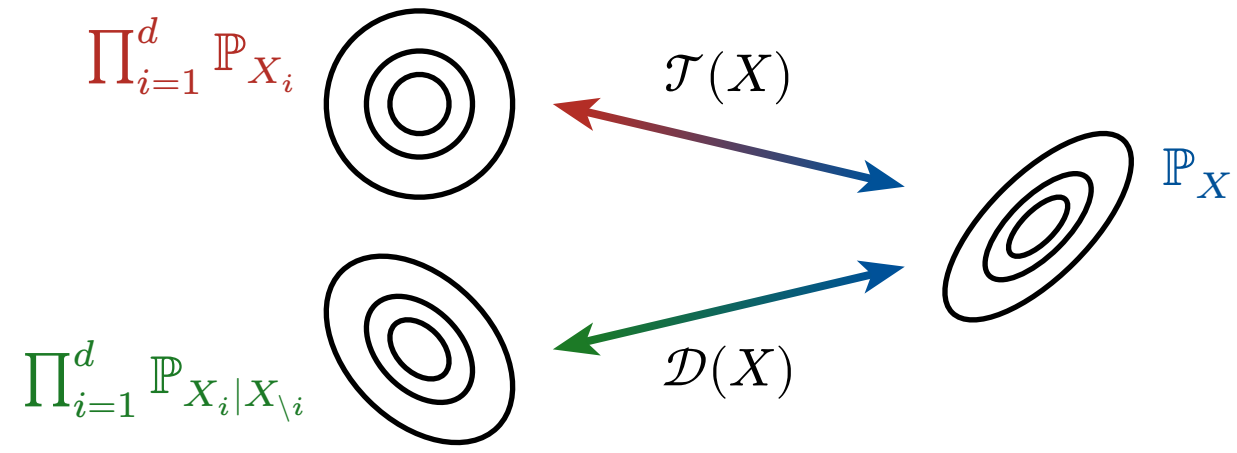

*Figure 2.* Flow Matching O-Information estimation (amortized).

Although MI serves as a fundamental measure for quantifying non-linear dependence between two random variables, its applicability becomes limited when analyzing systems comprising more than two variables, which is a common scenario e.g., in neuroscience (Ganmor et al., 2011) and climatology (Runge et al., 2019). A natural multivariate extension of MI is the interaction information, or co-information, defined via an inclusion-exclusion formula. While co-information can reflect the interplay between synergy and redundancy, its scalability and interpretability are limited for large systems. A more computable and interpretable generalization is provided by O-information (Bounoua et al., 2024; Rosas et al., 2019), defined as

$$\Omega(X) = \mathcal{T}(X) - \mathcal{D}(X),$$

where the redundancy term $\mathcal{T}(X)$ and the synergy term $\mathcal{D}(X)$ are given by

$$\mathcal{T}(X) = \sum_{i=1}^{d} \mathsf{h}(X_i) - \mathsf{h}(X),$$

$$\mathcal{D}(X) = \mathsf{h}(X) - \sum_{i=1}^{d} \mathsf{h}\Big(X_i \;\Big|\; X_{\setminus i}\Big).$$

Here $X_{\setminus i}$ denotes $(d-1)$-dimensional vector obtained by removing the $i$-th component from $X$. Intuitively, O-Information measures the overall strength and nature of interactions in a multivariate system. A positive $\Omega(X)$ indicates redundancy dominance, meaning that similar information is duplicated across different variables. Conversely, a negative $\Omega(X)$ corresponds to synergy, reflecting information that emerges only from the joint consideration of all variables, beyond what can be captured by pairwise interactions.

Although less straightforward, O-information also admits a DoE representation, enabling its estimation via our framework. To derive it, we consider a $d \times d$ matrix $\mathrm{M}$ of i.i.d. rows from $\mathbb{P}_X$. By construction,

$$\mathsf{h}(\operatorname{diag} \mathrm{M}) = \sum_{i=1}^{d} \mathsf{h}(X_i),$$

$$\mathsf{h}(\operatorname{diag} \mathrm{M} \mid \operatorname{offdiag} \mathrm{M}) = \sum_{i=1}^{d} \mathsf{h}\Big(X_i \;\Big|\; X_{\setminus i}\Big),$$

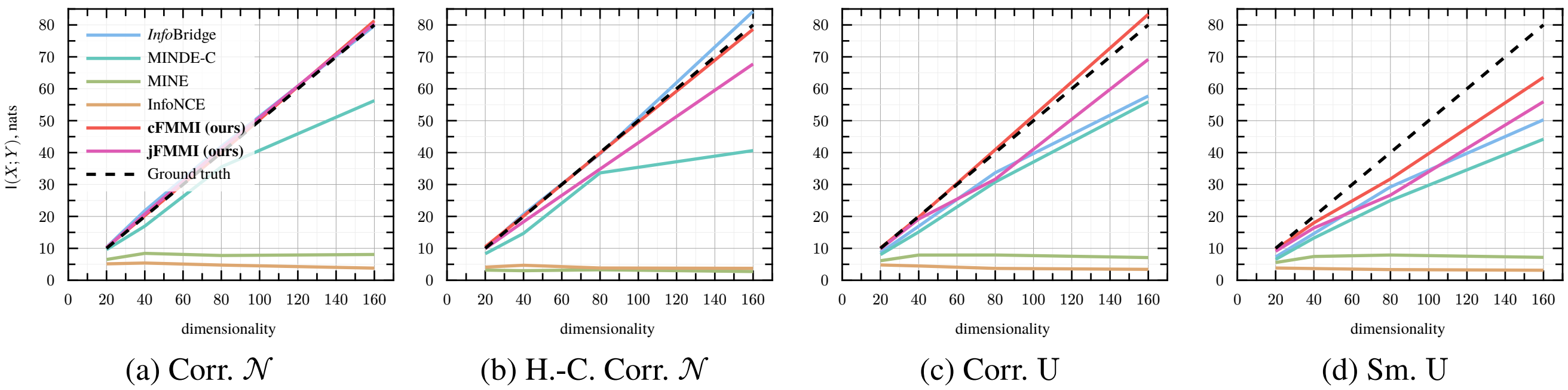

(a) Corr. $\mathcal{N}$ (b) H.-C. Corr. $\mathcal{N}$ (c) Corr. U (d) Sm. U

*Figure 3.* Comparison of MI estimates across dimensions and MI for high mutual information, adopted from (Kholkin et al., 2026).

Therefore, $\mathcal{T}(X) = \mathsf{h}(\operatorname{diag} \mathrm{M}) - \mathsf{h}(X)$ and $\mathcal{D}(X) = \mathsf{h}(X) - \mathsf{h}(\operatorname{diag} \mathrm{M} \mid \operatorname{offdiag} \mathrm{M})$.

While one can apply FMDoE to approximate each term individually, we suggest a more artful approach. Instead of learning two velocity fields, we employ a single $\hat{v}$ that takes two-dimensional $t \in \mathbb{R}^2$ as a time input. In this amortized setup, $t = (0, 0)$ corresponds to $\mathbb{P}_X$, $t = (1, 0)$ — to $\prod_{i=1}^{d} \mathbb{P}_{X_i}$, and $t = (0, 1)$ — to $\prod_{i=1}^{d} \mathbb{P}_{X_i \mid X_{\setminus i}}$. The regression objective (7) and affine paths (8) are extended trivially to this multidimensional time approach. We argue that this technique is also applicable to other information-theoretic quantities, e.g., to Transfer Entropy (Munoz et al., 2025).

## 7. Experiments

### 7.1. Mutual Information

To evaluate our estimator, we follow the experimental protocol from Section 5.4 in (Kholkin et al., 2026). We approximate $v$ via a two-layer MLP with hidden dimensionality 512. AdamW optimizer (Loshchilov and Hutter, 2019) with learning rate $10^{-3}$ is employed, $10^4$ gradient steps are used. Training set size is $10^5$ samples, test size is $10^4$. For details, please, refer to the supplementary source code.

The benchmark features *Correlated Normal*, *Half-cube Correlated Normal*, *Correlated Uniform* and *Smoothed Uniform* distributions from the `mutinfo` Python3 package (Butakov et al., n.d.). The results are provided in Figure 3. Note that, despite using $\times$ 20 fewer gradient steps, cFMMI consistently outperforms MINDE and *Info*Bridge — two state-of-the-art estimators for high-dimensional data. Compared to these methods, FMMI estimates were also more stable and did not require averaging over the last $n$ epochs.

### 7.2. O-Information

In this section, we validate our **FMOI** estimator on a set of synthetic becnhmarks and real data. For all experiments, we reuse the network from the previous section, but reduce the hidden dimensionality to 128. Other settings (optimizer, learning rate, etc.) are also the same. For more details, please, refer to the supplementary source code.

**Synthetic tests.** For the synthetic tests, we adopt a general Gaussian-based methodology from (Bounoua et al., 2024) but employ a more diverse set of covariance matrices. In particular, for each dimensionality, we use samples from the standard Wishart distribution: $\Sigma = \mathrm{A}^\mathsf{T}\mathrm{A}$, where $\mathrm{A} \sim \mathcal{N}(0, \mathrm{I} \otimes \mathrm{I})$. The following proposition provides a general expression for $\Omega(X)$ in the Gaussian case:

**Proposition 7.1.** For a $d$-dimensional Gaussian vector $X \sim \mathcal{N}(m, \Sigma)$, the O-information is given by

$$\Omega(X) = \frac{1}{2} \left( \sum_{i=1}^{d} \sigma_i^2 + (d-2) \log \det \Sigma - \sum_{i=1}^{d} \log \det \Sigma_{\setminus i} \right),$$

where $\Sigma_{\setminus i}$ is $\Sigma$ without its $i$-th row and column.

*Proof.* By the chain rule, $\mathsf{h}(X_i \mid X_{\setminus i}) = \mathsf{h}(X) - \mathsf{h}(X_{\setminus i})$, so the definition of O-information can be written as

$$\Omega(X) = \sum_{i=1}^{d} \mathsf{h}(X_i) + (d-2)\ \mathsf{h}(X) - \sum_{i=1}^{d} \mathsf{h}\big(X_{\setminus i}\big).$$

Substituting the entropies for a Gaussian vector yields

$$\Omega(X) = \frac{1}{2} \sum_{i=1}^{d} \log(2\pi e\, \sigma_i^2) + \frac{d-2}{2} \log\big((2\pi e)^d \det \Sigma\big) - \frac{1}{2} \sum_{i=1}^{d} \log\big((2\pi e)^{d-1} \det \Sigma_{\setminus i}\big),$$

which simplifies to the stated expression. □

For each $d$, we generate 20 covariance matrices. For each such matrix $\Sigma$, we sample $10^3$ train and test samples from $\mathcal{N}(0, \Sigma)$ and train FMOI for $10^4$ gradient steps in a single-batch regime. The results are presented in Table 1.

Since effective dimensionality of the problem is $d^2$ ($d$ for the state space and $d \times (d-1)$ for the conditions), $10^3$ samples are typically insufficient for complete probability distribution reconstruction in such a high-dimensional setup. Despite that, FMOI successfully recovers the true value of $\Omega(X)$ within a $10 - 15\%$ margin. We therefore believe that our method is suitable for the next task.

*Table 1.* FMOI results for synthetic Gaussian tests, $10^3$ samples, 20 runs per $d$.

| | dimensionality $d$ | | | | |
|---|---|---|---|---|---|
| Value | 3 | 4 | 5 | 6 | 8 |
| $\mathbb{E}\lvert\Omega(X)\rvert$ | 0.34 | 0.77 | 1.09 | 1.35 | 2.47 |
| $\mathbb{E}\big\lvert\hat{\Omega}(X) - \Omega(X)\big\rvert$ | 0.05 | 0.08 | 0.14 | 0.09 | 0.45 |
| $\dfrac{\mathbb{E}\big\lvert\hat{\Omega}(X) - \Omega(X)\big\rvert}{\mathbb{E}\lvert\Omega(X)\rvert}$ | 15% | 10% | 13% | 7% | 18% |
| $\mathbb{E}\,\dfrac{\big\lvert\hat{\Omega}(X) - \Omega(X)\big\rvert}{\lvert\Omega(X)\rvert}$ | 29% | 15% | 15% | 10% | 18% |

**fMRI data.** Functional magnetic resonance imaging (fMRI) is a non-invasive neuroimaging modality that measures brain activity indirectly via blood-oxygen-level-dependent (BOLD) signals, reflecting changes in local hemodynamics associated with neural activation. fMRI data are inherently high-dimensional, spatiotemporal, and noisy, consisting of time series of voxel-wise signals across the brain. The complex dependence structure across brain regions, subjects, and experimental conditions makes fMRI a natural application domain for information-theoretic analysis.

The fMRI data was obtained from the Human Connectome Project (HCP) 1200 Subjects Release (*1200 Subjects Data Release Reference Manual*, 2017). In this study, we rely on the preprocessed and curated dataset provided by (Kirova et al., 2025). The sample included 581 healthy, right-handed participants. Participants completed a series of task-based fMRI paradigms (Barch et al., 2013) designed to assess seven major cognitive domains, sampling the diversity of large-scale human brain networks: visual, motion, somatosensory, and motor systems; category specific representations; working memory/cognitive control systems; language processing (semantic and phonological); social cognition (Theory of Mind); relational processing; and emotion processing. A detailed description of all tasks is provided in Appendix C. Each participant performed seven tasks with two conditions, resulting in 14 distinct brain states (Table 2) and 8134 fMRI data units in total.

Kirova et al. (2025) identified unique sets of brain regions whose activity is most strongly associated with cognitive conditions, achieving high classification accuracy. Authors represented each fMRI data unit by the averaged activity values across brain regions. By focusing on the mean activity values, they aim to demonstrate that each brain state is associated with a unique network of regional activations, which can serve as a reliable signature for classification. In this work, we compute the O-Information specifically on these vectors of mean regional activity for each brain state. This allows us to quantify higher-order statistical dependencies among the selected brain regions and to characterize whether each cognitive state is encoded in a predominantly independent, redundant, or synergistic manner at the level of regional mean activations.

Kirova et al. (2025) suggest that for brain states with high classification accuracy (above 0.9), it is possible to identify a small set of unique brain regions whose mean activity is sufficient for reliable classification. In contrast, for the remaining states, the number of contributing regions is substantially larger and, in some cases, approaches coverage of most regions in the brain atlas, indicating more distributed neural representations. A detailed breakdown of the number of identified regions is provided in Appendix C.

We compute the O-information for all brain states, including both those characterized by a small number of identified regions and those involving a large, distributed set of regions. For each setup from the fMRI dataset, we perform 10 independent runs with different random seeds. In each run, the dataset of 581 samples is split into training (70%) and test sets (30%). Due to the limited dataset size, training employs a full-batch update and thus one gradient step per each of $10^4$ epochs. For each run, we compute the median O-information over the last 100 training epochs on the test set. The results are summarized in Figure 4 using box plots per region, where each box represents the distribution of the 10 seed-wise median values.

For states with a small number of significant regions, the O-information is consistently close to zero, being slightly positive in some cases and slightly negative in others. This behavior indicates that the selected regions encode largely independent or weakly interacting information, suggesting that the classifier relies on a compact set of complementary features with minimal redundancy or synergy. In this regime, near-zero O-information supports the interpretation that the identified regions are well-localized and capture distinct aspects of the underlying neural processes.

In contrast, for states associated with a large number of regions, we observe consistently high positive O-information. This pattern reflects strong higher-order statistical dependencies and redundancy across brain areas, indicating that information about these cognitive states is represented in a highly distributed and overlapping manner. Such elevated O-information is consistent with more global and integrative brain states, where multiple regions contribute correlated or redundant signals to the overall representation.

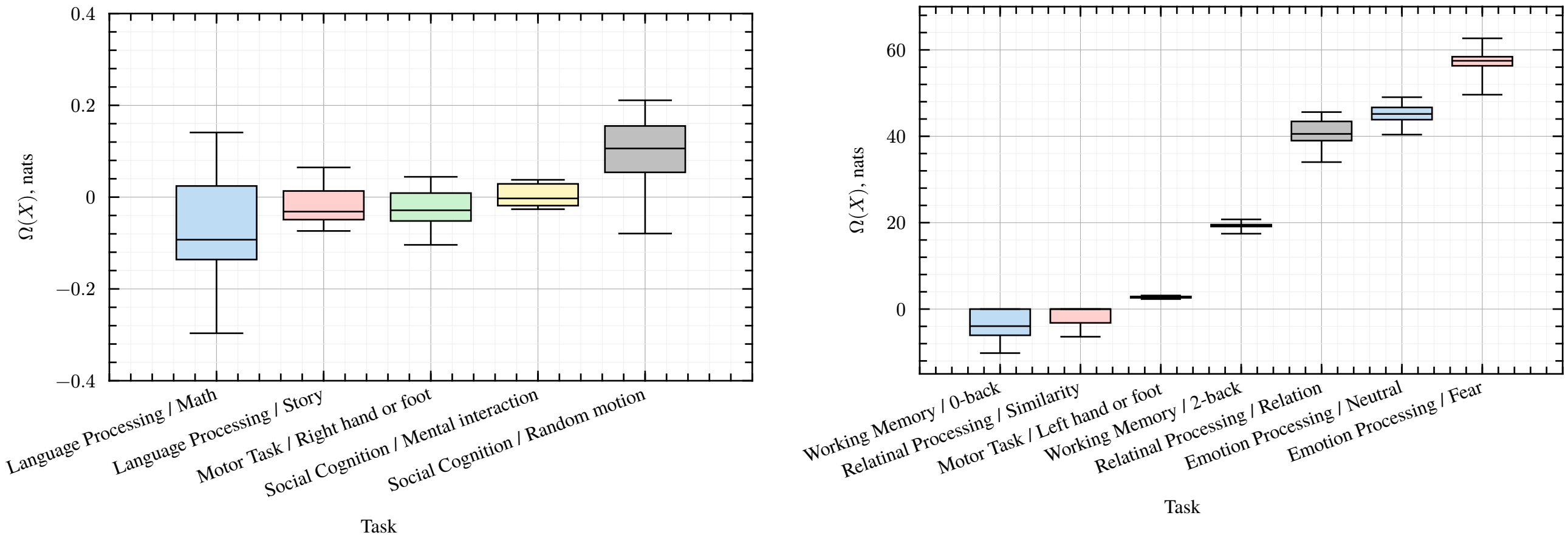

*Figure 4.* O-Information estimates for fMRI data (10 seeds): setups with low **(left)** and high **(right)** absolute value of O-Information. Low or even negative values indicate a successful selection of the most relevant regions; high positive values suggest that selected regions are redundant (encode similar information).

## 8. Discussion

To date, entropy and mutual information estimation remains extremely challenging, particularly for high-dimensional random vectors with complex dependencies. To address this, we introduce novel estimators based on flow matching — **FMDoE** and **FMMI** respectively. FMMI is a special case of FMDoE and comes in two variants: a conditional (**cFMMI**) and a joint (**jFMMI**) formulation.

In contrast to traditional generative MI estimators, our method constructs a flow between two data distributions, making it more akin to discriminative approaches. Evaluation on synthetic benchmarks demonstrates the superiority of the proposed method: while its computational load lies between that of discriminative and generative families, FMMI outperforms both in terms of accuracy, especially on high-dimensional and long-tailed data.

We provide theoretical guarantees proving our method's consistency. Furthermore, we propose Wasserstein-based variants of our estimators and establish a connection to FMDoE and FMMI.

Finally, a novel O-Information estimator (**FMOI**) is proposed. This method combines the FMDoE approach with an efficient multidimensional time amortization scheme, enabling fast training with only one network and two forward passes per batch. Applying FMOI to functional MRI data reveals redundancy in some groups of brain regions and a synergistic nature in others.

**Limitations.** Although FMMI is more lightweight than state-of-the-art diffusion-based MI estimators, it remains relatively demanding in terms of both computational and sample complexity.

**Reproducibility statement.** To ensure reproducibility of our results, we provide complete proofs in Appendix A and experimental details in Section 7. We also provide our source code in the supplementary material.

## A. Complete Proofs

**Lemma 4.1.** For any $t \in [0;1]$, let $X_t$ satisfy $\partial X_t / \partial t = v(X_t, t)$. Let $T \sim \mathrm{U}[0;1]$. Then

$$\mathsf{h}(X_1) - \mathsf{h}(X_0) = \mathbb{E} \operatorname{div} v(X_T, T)$$

*Proof of Lemma 4.1.* Let $x_t(x_0)$ be the solution to (4) corresponding to an initial condition $x_0$. Then

$$\begin{aligned}\mathsf{h}(X_1) - \mathsf{h}(X_0) &= \mathbb{E} \int_0^1 \operatorname{div} v(x_t(X_0), t)\, \mathrm{d}t = \int_0^1 \mathbb{E} \operatorname{div} v(x_t(X_0), t)\, \mathrm{d}t \\ &= \int_0^1 \mathbb{E}_{X_t \sim \mathbb{P}_{X_t}} \operatorname{div} v(X_t, t)\, \mathrm{d}t = \mathbb{E}_{T \sim \mathrm{U}[0;1], X_T \sim \mathbb{P}_{X_T}} \operatorname{div} v(X_T, T)\end{aligned}$$

□

**Theorem 4.2.** Consider $\mathbb{P}_{X,T}$ such that $p_t(x) \stackrel{\text{def}}{=} p(x \mid t)$ exists, is smooth, compactly supported, and $\log p_t$ is $L_t$-Lipschitz in $x$. Then, for any smooth vector field $\epsilon(x,t)$,

$$|\mathbb{E} \operatorname{div} \epsilon(X,T)| \leq \sqrt{\mathbb{E}\, L_T^2 \cdot \mathbb{E} \, \|\epsilon(X,T)\|_2^2}$$

*Proof of Theorem 4.2.* For any fixed $t$, define $S_t = \operatorname{supp} p_t$. Then

$$\mathbb{E}[\operatorname{div} \epsilon(X,T) \mid T = t] = \int_{S_t} p_t(x) \operatorname{div} \epsilon(x,t)\, \mathrm{d}x = \underbrace{\int_{S_t} \operatorname{div}(p_t(x)\, \epsilon(x,t))\, \mathrm{d}x}_{\substack{=0 \text{ by the divergence theorem} \\ \text{and vanishing property}}} - \int_{S_t} \langle \nabla p_t(x), \epsilon(x,t) \rangle\, \mathrm{d}x$$

Therefore,

$$\begin{aligned}|\mathbb{E} \operatorname{div} \epsilon(X,T)| &= \left| \mathbb{E} \int_{S_T} \langle \nabla p_T(x), \epsilon(x,T) \rangle\, \mathrm{d}x \right| = \left| \mathbb{E} \int_{S_T} \frac{p_T(x)}{p_T(x)} \langle \nabla p_T(x), \epsilon(x,T) \rangle\, \mathrm{d}x \right| \\ &= \left| \mathbb{E} \int_{S_T} p_T(x) \langle \nabla \log p_T(x), \epsilon(x,T) \rangle\, \mathrm{d}x \right| \\ &\qquad (\nabla \log f(x) = \nabla f(x) / f(x)) \\ &= \mathbb{E} \langle \nabla \log p_T(X_T), \epsilon(X_T, T) \rangle \leq \sqrt{\mathbb{E}\, \|\nabla \log p_T(X_T)\|_2^2} \cdot \sqrt{\mathbb{E}\, \|\epsilon(X_T,T)\|_2^2} \\ &\qquad (\text{Cauchy–Bunyakovsky–Schwarz}) \\ &\leq \sqrt{\mathbb{E}\, L_T^2} \cdot \sqrt{\mathbb{E}\, \|\epsilon(X_T,T)\|_2^2} = \sqrt{\mathbb{E}\, L_T^2 \cdot \mathbb{E}\, \|v(X_T,T) - \hat{v}(X_T,T)\|_2^2} \\ &\qquad (\log p_t \text{ is Lipschitz})\end{aligned}$$

□

**Theorem 4.4.** (FMDoE convergence rate) Under the assumptions of Corollary 4.3 and Theorem 4.4 in (Zhou and Liu, 2025), suppose $v$ is $K$-Lipschitz in $x$ and $\operatorname{var}[\operatorname{div} v(X_T, T)] \leq \sigma^2$. Let $\hat{v}$ be the FM estimate from $N$ i.i.d. samples from $\mathbb{P}_{X_0,X_1}$, and $\widehat{\mathrm{DoE}}_{\hat{v}}$ be the MC estimate of $\mathrm{DoE}_{\hat{v}}$. Then, $\forall \delta \in (0,1)$, with probability at least $1 - \frac{1}{N} - \delta$ over the random training and MC samples,

$$\left| \mathrm{DoE}_v - \widehat{\mathrm{DoE}}_{\hat{v}} \right| \leq \tilde{O}\left( \frac{\sqrt{\mathbb{E}\, L_T^2}\, K^{d/4}}{(1-T)^2 N^{1/(d+5)}} \right) + O\left( \frac{\sigma}{\sqrt{\delta N}} \right),$$

where $\tilde{O}$ ignores logarithmic factors in $d$, $\log N$, and $\log(1-T)$, and $f(x) = O(g(x))$ means $|f(x)| \leq |Cg(x)|$ for some $C > 0$.

*Proof of Theorem 4.4.* By the triangle inequality,

$$\left|\mathrm{DoE}_v - \widehat{\mathrm{DoE}}_{\hat{v}}\right| \leq \left|\mathrm{DoE}_v - \mathrm{DoE}_{\hat{v}}\right| + \left|\mathrm{DoE}_{\hat{v}} - \widehat{\mathrm{DoE}}_{\hat{v}}\right|. \tag{9}$$

Using Theorem 4.2, the first term is upper bounded by $\sqrt{\mathbb{E}\, L_T^2}\sqrt{\mathbb{E}\, \|v(X_T, T) - \hat{v}(X_T, T)\|_2^2}$. According to (Zhou and Liu, 2025, Theorem 4.4), with probability at least $1 - 1/N$,

$$\mathbb{E}_T \|v(X_T, T) - \hat{v}(X_T, T)\|_2^2 = \tilde{O}\left(\frac{K^{d/2}}{(1-T)^4 N^{2/(d+5)}}\right).$$

Let this high-probability event is denoted by $A$. Then, with probability at least $\mathbb{P}(A) \geq 1 - \delta$,

$$\left|\mathrm{DoE}_v - \mathrm{DoE}_{\hat{v}}\right| \leq \sqrt{\mathbb{E}[L_T^2]}\sqrt{\mathbb{E}_{X_T}\, \mathbb{E}_T \|v(X_T, T) - \hat{v}(X_T, T)\|_2^2} = \tilde{O}\left(\frac{\mathbb{E}[L_T^2]^{1/2} K^{d/4}}{(1-T)^2 N^{1/(d+5)}}\right).$$

Since $\widehat{\mathrm{DoE}}_{\hat{v}} = \frac{1}{N}\sum_{n=1}^N \operatorname{div} \hat{v}(x_T^n, T)$ is an unbiased estimator of $\mathrm{DoE}_{\hat{v}}$, with variance at most $\sigma^2/N$, the Chebyshev's inequality for any $\varepsilon > 0$ given training data gives

$$\mathbb{P}\left(\left|\mathrm{DoE}_{\hat{v}} - \widehat{\mathrm{DoE}}_{\hat{v}}\right| \geq \varepsilon\right) \leq \frac{\sigma^2}{\varepsilon^2 N}.$$

Setting $\varepsilon = \sigma/\sqrt{\delta N}$, the probability of failure is at most $\delta$. One can deduce that the total failure probability of both terms is at most $\delta + 1/N$, which completes the proof. □

**Corollary 4.6.** Suppose the assumptions of Theorem 4.2 hold and $L_t \leq L$ for some $L$ and every $t \in [0; 1]$. Then

$$|\mathsf{h}(X_1) - \mathsf{h}(X_0)| \leq L \cdot \mathsf{W}_2(\mathbb{P}_0, \mathbb{P}_1)$$

*Proof of Corollary 4.6.* By Lemma 4.1 and Theorem 4.2, one can note

$$|\mathsf{h}(X_1) - \mathsf{h}(X_0)| = |\mathbb{E} \operatorname{div} v(X_T, T)| \leq \sqrt{\mathbb{E}\, L_T^2}\sqrt{\mathbb{E}\, \|v(X_T, T)\|_2^2},$$

where $\sqrt{\mathbb{E}\, L_T^2} \leq L$ by assumption. Taking the infimum over $v \in V(\mathbb{P}_0, \mathbb{P}_1)$ and applying Lemma 4.5 for $p = 2$ completes the proof. □

## B. Relation to Other Estimators

In this section, we provide broader discussion on the connection of our method to the most relevant existing approaches.

**RFMI (Wang et al., 2025).** Flow-based mutual information estimators are numerous (Butakov et al., 2024b; Dahlke and Pacheco, 2025; Duong and Nguyen, 2023; Song and Ermon, 2020). However, only RFMI (Wang et al., 2025) and VCE (Chen et al., 2025) use flow matching as their backbone method.

Both techniques employ a Gaussian prior and two flow models to learn the data distribution: VCE estimates $\mathbb{P}_X$ and $\mathbb{P}_Y$, while RFMI estimates $\mathbb{P}_{X|Y}$ and $\mathbb{P}_X$. Since VCE uses the same copula trick as MIENF (Butakov et al., 2024b), it is inherently less flexible then RFMI.

In contrast to these methods, we propose learning a single flow from $\mathbb{P}_X \otimes \mathbb{P}_Y$ (or $\mathbb{P}_X$) to $\mathbb{P}_{X,Y}$ (or $\mathbb{P}_{X|Y}$). This approach avoids generative modeling altogether, focusing only on the coupling transform, which is sufficient for MI estimation (Chen et al., 2025).

**DRE-∞ (Choi et al., 2022).** Traditional discriminative MI estimators often fail in high-MI scenarios, a problem attributed to the "density chasm." This issue can be partially mitigated by using a telescoping chain of density ratio estimators (Rhodes et al., 2020). DRE-∞ advances this approach by learning a continuous chain of infinitesimal classifiers (Choi et al., 2022).

Similar to our method, DRE-∞ requires a probability path $\mathbb{P}_t$ between two distributions. However, it also demands that

$$\lim_{\Delta t \to 0} \frac{\mathrm{d}\,\mathbb{P}_{t+\Delta t}}{\mathrm{d}\,\mathbb{P}_t}(x)$$

is non-degenerate for any $x$ and $t$, a condition that is difficult to ensure in practice. Because our method is not a density ratio estimator, it is free from this limitation.

**"Loss Comparison" (Covert et al., 2020).** The connection between MI and traditional losses (e.g., MSE, MAE, accuracy) is well-established: if $X$ enables an accurate prediction of $Y$, the mutual information $\mathsf{I}(X;Y)$ is typically high (Cover and Thomas, 2006). However, this approach only provides crude bounds tied to a specific predictive model $\mathbb{P}_{Y|X}$ (Covert et al., 2020).

In contrast, our method does not merely predict $Y$ from $X$; it fully reconstructs the conditional distribution $\mathbb{P}_{Y|X}$, thereby achieving a far more accurate MI estimate.

## C. Application to fMRI data

Each fMRI data unit is represented as an activity matrix $\boldsymbol{X}$ of size $r \times l$, where each line $\boldsymbol{x}_i = (x_{i1}, ..., x_{il})$ corresponds to the activity of the $i$-th brain region over a time interval of length $l$. Each element $x_{i,j}$ of the activity matrix $\boldsymbol{X}$ corresponds to the activity level of the i-th brain region at time point $j$.

In (Kirova et al., 2025), participants were selected using three inclusion criteria to ensure high data quality. First, only individuals who successfully completed all seven HCP task paradigms were retained. Second, all cases flagged with quality issues during the HCP quality control process were excluded, retaining only participants with no reported imaging problems. Third, only adults aged 22 to 35 years at the time of scanning were included.

The final sample included 581 healthy participants, all right-handed. Participants were met with a series of tasks (Barch et al., 2013) designed to assess seven main cognitive domains sampling the diversity of human neural networks, including: 1) visual, motion, somatosensory, and motor systems; 2) category specific representations; 3) working memory/cognitive control systems; 4) language processing (semantic and phonological processing); 5) social cognition (Theory of Mind); 6) relational processing; and 7) emotion processing. Three of these tasks were completed in one session, while the remaining four were performed in another session. Below, we provide a brief description of each task.

**Working Memory.** This task was designed to study short-term memory and information retention processes. Participants were shown blocks of images depicting places, tools, faces, and body parts. Each run included blocks from four types of stimuli, with half of the blocks performed with a 2-back task requiring memory retention, and the other half with a 0-back task for comparison. The 2-back task required participants to remember the sequence of images and determine whether the current image matched the one shown two steps earlier. The 0-back task only required participants to identify whether the current image matched a given target. Task blocks alternated with fixation blocks, where participants looked at a cross on the screen, allowing researchers to track brain activity changes in the absence of cognitive load.

**Gambling Task.** This task simulated decision-making processes under uncertainty and risk. Participants played a card game in which they had to guess whether the number on a hidden card was higher or lower than 5. Depending on their answer, they could win (green upward arrow with "1"), lose (red downward arrow with "0.50"), or receive no reward (neutral outcome). The task was divided into blocks dominated by either winning or losing outcomes, enabling researchers to examine how the brain responds to reward anticipation and loss.

**Motor Task.** This task was aimed at mapping motor areas of the brain. Participants were presented with visual cues instructing them to perform movements such as tapping their fingers on the left or right hand, squeezing their toes, or moving their tongue. Movement blocks lasted 12 seconds, and each of the two runs contained 13 blocks, including tongue, hand, and foot movements, as well as fixation blocks. These tasks help identify motor cortex areas activated in response to different types of movements.

**Language Processing.** This task aimed to study language comprehension and arithmetic processing. In the story blocks, participants listened to short narratives, such as adaptations of Aesop's fables, and then answered questions about the story content. In the math blocks, participants solved verbal arithmetic problems. The tasks alternated, and the block durations were adjusted to ensure equal completion times, allowing comparison of brain activity during language processing and numerical operations.

**Social Cognition (Theory of Mind).** Participants were shown short video clips where geometric shapes (squares, circles, triangles) either interacted with each other or moved randomly. After each video, participants evaluated whether the shapes had intentions and were interacting or whether their movements were random. This task allowed researchers to study brain areas associated with understanding social interactions and recognizing intentions.

**Relational Processing.** This task involved analyzing relationships between objects on the screen. In one condition, participants determined which characteristic (shape or texture) differentiated object pairs and assessed whether this distinction applied to another pair. In the control condition, they simply identified whether the lower object matched one of the upper objects based on a given criterion. This task helped researchers study the brain's ability to perform comparative analysis and identify relationships between objects.

**Emotion Processing.** Participants were shown faces displaying fear or anger, as well as neutral figures, and had to choose one of two faces or figures that matched the presented stimulus. Task blocks alternated with fixation blocks, allowing researchers to assess brain activity during emotional stimulus processing compared to neutral conditions.

In the work (Kirova et al., 2025), authors represented each activity matrix $\boldsymbol{X}$ by the averaged activity values across brain regions. For each brain region $i = (1, ..., r)$ they calculated the mean value $\overline{\boldsymbol{x}}_i$ of the time series $\boldsymbol{x}_i$. Thus, for each activity matrix $\boldsymbol{X}$, we obtain a vector of mean activity values across all brain regions, which will be used in further analysis: $\boldsymbol{x}_{\text{mean}} = (\overline{\boldsymbol{x}}_1, ..., \overline{\boldsymbol{x}}_r)$.

The authors placed particular emphasis on the high classification accuracy achieved for the following brain states: *left hand or foot, right hand or foot, math, story, mental interaction, and random motion*. For these states, a relatively small number of identified brain regions is sufficient to achieve high classification performance based on their mean activity values. Moreover, the sets of regions are state-specific, forming compact and largely non-overlapping signatures for each cognitive condition. The HCP indices of the brain regions that are most significant for classification, together with their anatomical labels, are reported in Table 3.

As also demonstrated by the authors, for the remaining cognitive states the number of regions contributing to classification is substantially larger, approaching coverage of most regions in the brain atlas. This indicates more distributed and less localized neural representations, consistent with more globally integrated cognitive processes.

*Table 2.* Two brain states for each cognitive task, between which classification is made

| **Cognitive Task** | **State name 1** | **State name 2** |
|---|---|---|
| Working Memory | 0-back | 2-back |
| Gambling | Win | Loss |
| Motor Task | Left hand or foot | Right hand or foot |
| Language Processing | Story | Math |
| Social Cognition | Random motion | Mental interaction |
| Relational Processing | Relation | Similarity |
| Emotion Processing | Neutral | Fear |

*Table 3.* The HCP indices and anatomical labels of significant brain regions

| **Index** | **Label** |
|---|---|
| **Language Processing: Story** | |
| 80 | IFJa_L, IFJa, Fr, Inferior Frontal |
| 117 | AIP_L, AIP, Par, Superior Parietal |
| 150 | PGi_L, PGi, Par, Inferior Parietal |
| **Language Processing: Math** | |
| 44 | 6ma_L, 6ma, Fr, Paracentral Lobular and Mid Cingulate |

| **Language Processing: Math** | |
|---|---|
| 128 | STSda_L, STSda, Temp, Auditory Association |
| 135 | TF_L, TF, Temp, Medial Temporal |
| 137 | PHT_L, PHT, Temp, Lateral Temporal |
| 146 | IP0_L, IP1, Par, Inferior Parietal |
| 186 | V4_R, V4, Occ, Early Visual |
| 313 | TE1p_R, TE1p, Temp, Lateral Temporal |
| 337 | FST_R, FST, Occ, MT+Complex and Neighboring Visual Areas |
| 367 | LP_L, Thalamus: Lateral Posterior |
| **Social Cognition: Mental Interaction** | |
| 4 | V2_L, V2, Occ, Dorsal Stream Visual |
| 338 | V3CD_R, V3CD, Occ, MT+Complex and Neighboring Visual Areas |
| **Social Cognition: Random Motion** | |
| 13 | V3A_L, V3A, Occ, Dorsal Stream Visual |
| 128 | STSda_R, STSda, Temp, Auditory Association |
| 146 | IP0_L, IP0, Par, Inferior Parietal |
| 185 | V3_R, V3, Occ, Early Visual |
| 203 | MT_R, MT, Occ, MT+Complex and Neighboring Visual Areas |
| 309 | STSdp_R, STSdp, Temp, Auditory Association |
| **Motor Task: Right Hand or Foot** | |
| 39 | 5L_L, 5L, Par, Paracentral Lobular and Mid Cingulate |
| 55 | 6mp_L, 6mp, Fr, Paracentral Lobular and Mid Cingulate |
| 231 | 1_R, 1, Par, Somatosensory and Motor |
| **Motor Task: Left Hand or Foot** | |
| 8 | 4_L, 4, Fr, Somatosensory and Motor |
| 36 | 5m_L, 5m, Par, Paracentral Lobular and Mid Cingulate |
| 39 | 5L_L, 5L, Par, Paracentral Lobular and Mid Cingulate |
| 54 | 6d_L, 6d, Fr, Premotor |
| 55 | 6mp_L, 6mp, Fr, Paracentral Lobular and Mid Cingulate |
| 105 | PFcm_L, PFcm, Par, Early Auditory |
| 148 | PF_L, PF, Par, Inferior Parietal |
| 220 | 24dd_R, 24dd, Fr, Paracentral Lobular and Mid Cingulate |
| 231 | 1_R, 1, Par, Somatosensory and Motor |
| 232 | 2_R, 2, Par, Somatosensory and Motor |
| 281 | OP1_R, OP1, Par, Posterior Opercular |
| 285 | PFcm_R, PFcm, Par, Early Auditory |
| 296 | PFt_R, PFt, Par, Inferior Parietal |